# Warm-Start Flow Matching for Guaranteed Fast Text/Image Generation

**Minyoung Kim** [1]

## Abstract

Current auto-regressive (AR) LLMs, diffusion-based text/image generative models, and recent flow matching (FM) algorithms are capable of generating premium quality text/image samples. However, the inference or sample generation in these models is often very time-consuming and computationally demanding, mainly due to large numbers of function evaluations corresponding to the lengths of tokens or the numbers of diffusion steps. This also necessitates heavy GPU resources, time, and electricity. In this work we propose a novel solution to reduce the sample generation time of flow matching algorithms by a guaranteed speed-up factor, without sacrificing the quality of the generated samples. Our key idea is to utilize computationally lightweight generative models whose generation time is negligible compared to that of the target AR/FM models. The draft samples from a lightweight model, whose quality is not satisfactory but fast to generate, are regarded as an initial distribution for a FM algorithm. Unlike conventional usage of FM that takes a pure noise (e.g., Gaussian or uniform) initial distribution, the draft samples are already of decent quality, so we can set the starting time to be closer to the end time rather than 0 in the pure noise FM case. This will significantly reduce the number of time steps to reach the target data distribution, and the speed-up factor is guaranteed. Our idea, dubbed *Warm-Start FM* or WS-FM, can essentially be seen as a *learning-to-refine* generative model from low-quality draft samples to high-quality samples. As a proof of concept, we demonstrate the idea on some synthetic toy data as well as real-world text and image generation tasks, illustrating that our idea offers guaranteed speed-up in sample generation without sacrificing the quality of the generated samples.

## 1. Introduction

In this paper we deal with generative models for domains such as natural language texts or images. Despite plethora of recent literature on deep generative models, most notably auto-regressive (AR) LLMs and diffusion generative models, we particularly focus on the flow matching (FM) methods (Lipman et al., 2023a;b; Kim, 2025; Gat et al., 2024; Campbell et al., 2024). Our goal is to have a guaranteed speed-up factor for sample generation compared to existing FM methods without sacrificing the sample quality.

Given two distributions whose samples are accessible, FM aims to learn a generator (either in the form of diffusion or continuous-time Markov chain) that transports one distribution to the other. Most FM models deal with pure noise (e.g., Gaussian white noise or uniform) as the initial distribution and the data samples that we aims to mimic as the target distribution. The generation of samples from FM models often requires many function evaluations or time steps to enable accurate transportation from noise to data, leading to a time-consuming and computationally demanding generation (inference) stage.

Our key insight is that this slow generation or a large number of time steps originates from too naive initial distribution that has zero information about the target distribution. Since we already have access to the target data samples, we can first train some computationally lightweight generative models (e.g., small LSTM networks or GAN models). The generation time for these lightweight models is negligible compared to that of the target FM models. Our main idea is that the draft samples from a lightweight model, whose quality is not satisfactory but fast to generate, are then regarded as an initial distribution for a FM algorithm. Unlike conventional usage of FM that takes a pure noise initial distribution, the draft samples are already of decent quality, so we can set the starting time to be closer to the end time rather than 0 in the pure noise FM case. This will significantly reduce the number of time steps to reach the target data distribution, and even further the speed-up factor is guaranteed. Our idea, dubbed *Warm-Start FM* or WS-FM, can essentially be seen as a *learning-to-refine* generative model from low-quality draft samples to high quality samples.

Following the conventions in diffusion models and flow matching algorithms, we let $P_1(x)$ be the target data dis-

[1]Samsung AI Center Cambridge, United Kingdom. Correspondence to: Minyoung Kim <mikim21@gmail.com>.

*Preprint.*

Warm-Start Flow Matching for Guaranteed Fast Text/Image Generation

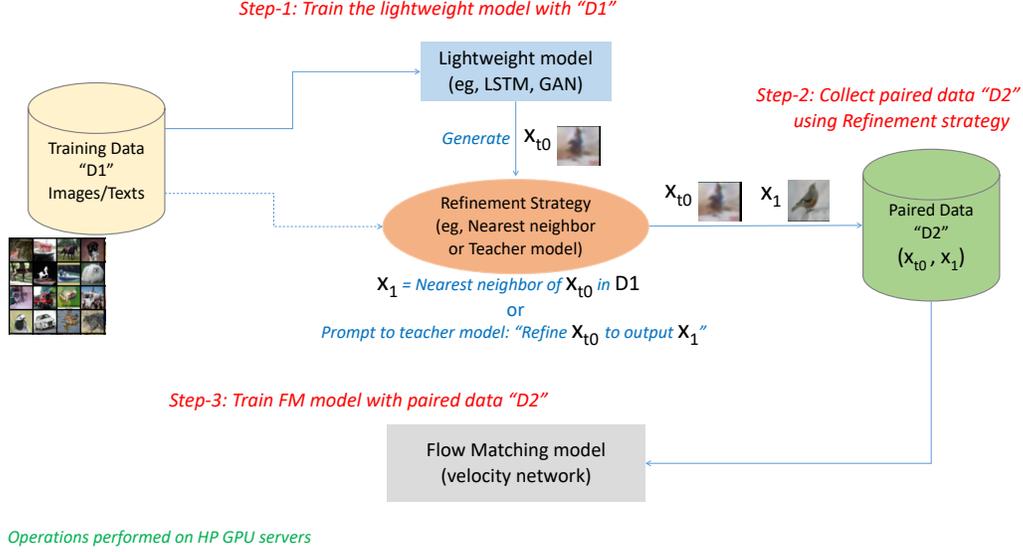

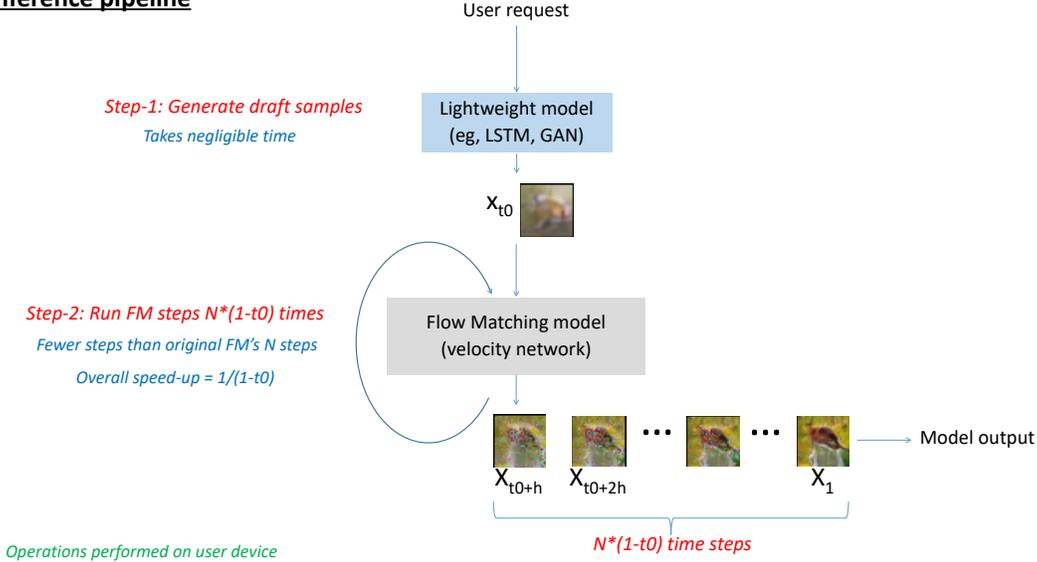

*Figure 1.* Training and inference pipelines for our warm-start flow matching method.

tribution, and $P_0(x)$ be the pure noise distribution (e.g., Gaussian white noise for continuous domain and uniform or mask-delta[1] distributions for discrete domain). Conceptually the flow matching (FM) algorithms (Lipman et al., 2023a; Albergo & Vanden-Eijnden, 2023; Kim, 2025) regard $P_0$ as the initial distribution at time $t = 0$, and $P_1$ as the destination distribution at time $t = 1$, and consider interpolation $x_t$ at $t \in [0, 1]$ between two end points $x_0 \sim P_0$ and $x_1 \sim P_1$. The simplest form is the linear interpolation: $x_t = tx_1 + (1 - t)x_0$ in the state space (e.g., in continuous domain) or $x_t \sim tP_1(x) + (1 - t)P_0(x)$ in the probability space (e.g., in discrete domain). This interpolation naturally forms a velocity vector field via time derivatives, and the FM training amounts to learning the velocity field via a neural network $u(t, x_t)$. Once trained, the learned velocity field dictates the transportation from $P_0$ to $P_1$, thereby generating samples from $P_1$.

Although diffusion generative models (Sohl-Dickstein et al., 2015; Ho et al., 2020; Song et al., 2021) are relatively more popular, there are two main reasons why we opt for FM in our work. First, diffusion models aim to revert a noising process, and the score estimation required in the reverse process is often more difficult to learn than velocity learning in FM. Secondly, since diffusion models rely on the noising process to convert the data distribution to pure noise, the resulting end distribution that serves as an initial distribution for generation, is quite restricted to a specific noise distribution (e.g., Gaussian white noise or the mask-delta). On the other hand, FM has more freedom in choosing the initial distribution.

---
[1] In many discrete flow matching and diffusion models, e.g. (Gat et al., 2024; Campbell et al., 2024), $P_0(x) = \delta_{mask}(x)$ is often used where $mask$ is a special state apart from normal states.

FM is more attractive in this regard, but conventional practice in FM incurs some other issues. First, the sample generation with FM usually takes long due to many time steps or function (neural network) evaluations for accurate transportation. Secondly, the pairing (or coupling) between $P_0$ and $P_1$ for interpolation is usually inefficient – In practice the independent pairing $Q(x_0, x_1) = P_0(x_0) \cdot P_1(x_1)$ is often used. For high state dimensionality, however, such an independent coupling can makes the velocity estimation highly inefficient, which makes FM training difficult. Some more informative pairing can make the training more efficient (Shi et al., 2023; Kim, 2025), however, coming up with an effective non-independent pairing is often difficult in practice.

## 2. Background on Discrete Flow Matching

Even though our idea can be straightforwardly applied to continuous-state flow matching algorithms, we focus on discrete flow matching in our paper due to its wide applicability to both language and vision domains. This section reviews some technical background on discrete flow matching.

For notations, we let $x = (x^1, \ldots, x^N)$ be the state that consists of $N$ tokens, each of which takes a value from $[V]$ (i.e., $x^i \in \{1, \ldots, V\}$). We build a diffusion generative model in the continuous time $t \in [0, 1]$, and $x_t = (x_t^1, \ldots, x_t^N)$ represents the state at time $t$. We denote by $P_1(x)$ the data distribution that we aim to learn where the subscript 1 comes from the convention where the model distribution of terminal state $x_1$ is supposed to match the target distribution. Accordingly the initial distribution is denoted as $P_0(x)$ which is typically a pure noise distribution (e.g., uniform over $[V]^N$ or a delta function at a special token, often called the mask token, in which case we extend $[V]$ to include it).

Discrete flow models (Campbell et al., 2024; Gat et al., 2024) deal with the continuous-time Markov chain (CTMC) model, and we denote the kernel for the $i$-th token by $u_t^i(x_t, x_{t+h}^i)$, where the probability of $x_{t+h}^i$ is given as:

$$\delta_{x_t^i(\cdot)} + h u_t^i(x_t, \cdot) \quad (1)$$

where $h$ is small positive, ideally infinitesimal. For the full distribution for $x_{t+h}$ over $N$ tokens, we assume token-wise independence that is adopted by most discrete generative models to circumvent exponentially many configurations, otherwise computationally infeasible.

The discrete flow matching (DFM) (Campbell et al., 2024; Gat et al., 2024) conforms to the general flow matching framework (e.g., see (Lipman et al., 2023b) or the unified bridge framework described in (Kim, 2025)). The main idea, regardless of continuous or discrete state space, is to come up with the so called *coupling distribution* $Q(x_0, x_1)$ and *pinned marginals* $P_t(x_t|x_0, x_1)$ and its generators $u_t(x_t, \cdot|x_0, x_1)$ which are also pinned at $x_0$ and $x_1$. Once these pinned marginal generators are learned, one can have (unconditional) generators $u_t(x_t, \cdot)$ that are proven to produce samples that follow the marginal distributions $P_t(x_t)$ that are induced from $Q(x_0, x_1)$ and $P_t(x_t|x_0, x_1)$. Hence, if the coupling $Q$ satisfies the marginal constraint $Q(x_1) = P_1(x_1)$, then we can guarantee that the samples $x_1$ from the generator $u_t(x_t, \cdot)$ follow the target $P_1$.

We mainly follow the derivations in (Gat et al., 2024), and specifically, we adopt the pinned generators

$$u_t^i(x_t, \cdot|x_0, x_1) = \sum_{j=1}^{J} a_t^{i,j} w^{i,j}(\cdot|x_0, x_1) + b_t^i \delta_{x_t^i}(\cdot) \quad (2)$$

where $a_t^{i,j} = \ldots$, $b_t^i = \ldots$, and $\{\kappa_t^{i,j}\}_{j=1}^{J}$ forms convex combinations (i.e., non-negative and summed to 1). It can be shown that we then have the pinned marginals,

$$P_t^i(x_t^i|x_0, x_1) = \sum_{j=1}^{J} \kappa_t^{i,j} w^{i,j}(x_t^i|x_0, x_1) \quad (3)$$

Typical choices for the distributions $w^{i,j}(x_t^i|x_0, x_1)$ are simple delta functions, e.g., $\delta_{x_0^i}(\cdot)$ and $\delta_{x_1^i}(\cdot)$, or the uniform distribution. By defining the marginals as

$$P_t(x) = \sum_{x_0, x_1} Q(x_0, x_1) P_t(x_t|x_0, x_1) \quad (4)$$

and recall that $P_t(x_t|x_0, x_1)$ is defined as the product of (3) from the token-wise independence assumption. From the flow matching principle (Lipman et al., 2023b; Kim, 2025), the unconditional generator defined as

$$u_t^i(x_t, \cdot) = \frac{1}{P_t(x_t)} \mathbb{E}_Q[u_t^i(x_t, \cdot|x_0, x_1) P_t(x_t|x_0, x_1)] \quad (5)$$

admits $P_t(x_t)$ as its marginal distributions.

However, since it is generally not tractable to compute (5) due to the difficult marginalization, we instead estimate it with a neural network by optimizing the following:

$$\min_\theta -\sum_{i,j} \mathbb{E}_{t, x_0, x_1} \mathbb{E}_{w^{i,j}(c|x_0, x_1)}[\log v_\theta(t, x_t)_{i,j,c}] \quad (6)$$

where $v_\theta(t, x_t)$ is the neural network approximator that outputs $(N \times J \times V)$ tensor, and the first expectation is with respect to the uniform over $t \in [0, 1]$, $Q(x_0, x_1)$, and

$P_t(x_t|x_0, x_1)$. At the sampling time, we generate samples from the model using

$$u_t^i(x_t, \cdot) = \sum_{j=1}^{J} a_t^{i,j} v_\theta(t, x_t)_{i,j} + b_t^i \delta_{x_t^i}(\cdot) \qquad (7)$$

## 3. Our Method

We aim to address the above-mentioned two issues that appear in the conventional FM practice.

First, for the initial distribution of FM, instead of pure noise, we adopt a lightweight fast generative models such as small-sized LSTM (Hochreiter & Schmidhuber, 1997) for discrete sequences or GAN (Radford et al., 2015) for images that is trained on the data samples. Since the samples from such a lightweight model are already closer to the target $P_1(x)$ than pure noise, we can set the start time in the flow matching (denoted by $t_0$) to be closer to final time $t = 1$, e.g., $t_0 = 0.8$, instead of original FM's $t_0 = 0$. The (initial) distribution of the samples generated by the lightweight model is denoted as $P_{t_0}(x)$, and we build a FM bridge between $P_{t_0}$ at $t = t_0$ and $P_1$ at $t = 1$. We call this idea *Warm-Start Flow Matching* (WS-FM) contrast to original FM's cold start initial. Since sample generation from $P_{t_0}(x)$ is negligibly fast, the number of WS-FM time steps is *guaranteed* to be reduced from $N$ to $N \cdot (1 - t_0)$ where $N$ is the time steps required in the original FM. The speed-up in data generation is then $1/(1 - t_0)$ (e.g., $\times 5$ speed-up if we use $t_0 = 0.8$). The hyperparameter $t_0$ in our WS-FM is chosen empirically by seeking for the largest $t_0$ that does not degrade the final sample quality.

Secondly, for more efficient pairing beyond the conventional independent pairing, we employ the concept of *refinement*: an initial sample $x_{t_0} \sim P_{t_0}(x)$ is *refined* to a sample $x_1$ so that $x_1$ (approximately) follows $P_1(x)$. To exemplify it in the English text generation cases where $P_1(x)$ represents the set of natural English texts, we may use off-the-shelf LLMs for refinement. For a draft text $x_{t_0} \sim P_{t_0}(x)$, for instance, we ask LLMs to respond with the following prompt:

> *"Refine the input text ${x_{t_0}}$ so that it becomes a more natural and grammatically correct English text, but not too different from the input text."*

Alternatively, especially in the non-text domains or if strong foundation models like LLMs are not available, the refinement of $x_{t_0}$ can be done by finding the nearest neighbors of $x_{t_0}$ in the dataset of $P_1(x)$. We indeed use this strategy in our image generation experiments.

Formally, the refinement can be viewed as having a coupling distribution $Q(x_{t_0}, x_1) = P_{t_0}(x_{t_0}) \cdot P_{refine}(x_1|x_{t_0})$ [2]

---
[2] In theory, the flow matching requires the coupling $Q(x_{t_0}, x_1)$

```
for _ in range(niters):

    # samples from data and noise distributions
    x_1 ~ Data
    x_0 ~ Uniform noise

    # sample time
    t ~ Uniform(0, 1)

    # sample x_0-x_1 interpolated points at t
    x_t = Interp(t, x_1, t1 = 1, x_0, t0 = 0)

    logits = DFNet(x_t, t)  # u(t, x_t)
    loss = CrossEntropy(logits, x_1)
    # Update DFNet() with loss
```

```
t0 = 0.8  # warm starting time (0.8 -> x5 speed-up)
for _ in range(niters):

    # sample (candidate, refined) pairs
    (x_t0, x_1) ~ (Small/fast model generated, Refined)

    # sample time
    t ~ Uniform(t0, 1)

    # sample x_t0-x_1 interpolated points at t
    x_t = Interp(t, x_1, t1 = 1, x_t0, t0 = t0)

    logits = DFNet(x_t, t)  # u(t, x_t)
    loss = CrossEntropy(logits, x_1)
    # Update DFNet() with loss
```

*Figure 2.* Training algorithm comparison between DFM (Gat et al., 2024) (LEFT) and our WS-DFM (RIGHT). Changes shown as red.

```
t = 0.0  # start time = 0.0
x_t ~ Initial as uniform noise from vocabulary
while t < 1.0:
    p1 = softmax(DFNet(x_t, t))
    u = (p1 - x_t) / (1.0 - t)
    x_t ~ Categorical(x_t + h * u)
    t += h

# return x_t as a generated sample
```

```
t = t0  # start time = t0 (= 0.8)
x_t ~ small/fast model generated rough sentences
while t < 1.0:
    p1 = softmax(DFNet(x_t, t))
    u = (1.0-t0) * (p1 - x_t) / (1.0 - t)  # time-warping
    x_t ~ Categorical(x_t + h * u)
    t += h

# return x_t as a generated sample
```

*Figure 3.* Inference algorithm comparison between DFM (Gat et al., 2024) (LEFT) and our WS-DFM (RIGHT). Changes as red.

where $P_{refine}$ is one of the mapping/refinement strategies that we discussed. We collect paired samples $(x_{t_0}, x_1)$ from $Q(x_{t_0}, x_1)$, and they serve as training data for FM training. During training the interpolation is confined to $t \in [t_0, 1]$ where the linear interpolation takes the form $x_t = \frac{1-t}{1-t_0} x_{t_0} + t x_1$. Training amounts to minimizing discrepancy between $u(t, x_t)$ and the velocity at $x_t$. At the inference stage, we can use the learned velocity field $u(t, x_t)$ starting from $t = t_0$ to gradually transport $P_{t_0}$ to $P_1$. As an inverse operation for our modified interpolation, the velocity *time-warping* may be needed: $u \leftarrow \frac{1-t_0}{1-t} u$ during generation steps.

We visualize the overall training and inference pipelines for our WS-FM in Fig 1. For the specific discrete-domain DFM (Gat et al., 2024), we differentiate algorithms (pseudocodes) between DFM and our WS-DFM in Fig. 2 (training) and Fig. 3 (inference/generation) where we highlight the main changes.

## 4. Experiments

For the proof-of-concept empirical verification of the proposed idea, we demonstrate several experimental results. For our experiments we predominantly focus on discrete flow matching (DFM) algorithms (Gat et al., 2024; Campbell et al., 2024) which are advantageous over the continu-satisfies the boundary marginal constraints, namely $Q(x_{t_0}) = P_{t_0}(x_{t_0})$ and $Q(x_1) = P_1(x_1)$. The former is easily met by construction, while the latter is often not strictly satisfied. As a remedy, we may modify $P_{refine}(x_1|x_{t_0})$ as a mixture of $P_1(x_1)$ and the refinement mapping, which can be done by randomly injecting $x_1 \sim P_1(x_1)$ samples to the paired data, as we did in our experiments in Sec. 4.3.

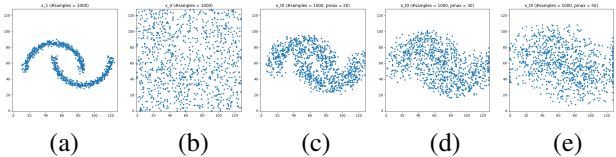

| (a) | (b) | (c) | (d) | (e) |

*Figure 4.* (Two moons) (a) Target distribution $P_1(x)$. (b) Pure uniform noise distribution $P_0(x)$ used in DFM (Gat et al., 2024). (c–e) Three lightweight draft models adopted in our WS-DFM: (c) = pretty good, (d) = fair, (e) = poor.

*Table 1.* (Two moons) Sample generation results. The symmetric KL (SKL) divergences between generated and true samples, and the number of function evaluations (NFE) are shown. In our WS-DFM, those cases where the sample quality is no worse than original DFM are marked as ✔.

|  | $t_0$ | SKL (↓) | NFE (↓) |
|---|---|---|---|
| Original DFM (Gat et al., 2024) | 0 | 0.62 | 20 |
| WS-DFM + Pretty good draft model | 0.95 | 0.74 ✘ | 1 |
|  | 0.9 | 0.54 ✔ | 2 |
|  | 0.8 | 0.37 ✔ | 4 |
| WS-DFM + Fair draft model | 0.8 | 0.86 ✘ | 4 |
|  | 0.5 | 0.51 ✔ | 10 |
| WS-DFM + Poor draft model | 0.8 | 1.35 ✘ | 4 |
|  | 0.5 | 0.64 ✘ | 10 |
|  | 0.35 | 0.54 ✔ | 13 |

ous counterparts in that the text/language data can be easily handled. We denote by $P_1(x)$ the data distribution, and the initial distribution is denoted as $P_0(x)$ which is typically a pure noise distribution (e.g., uniform over $[V]^N$ or a delta function at a special token, often called the mask token, in which case we extend $[V]$ to include it). Unlike this original treatment, we have the initial distribution of draft samples $x_{t_0} \sim P_{t_0}(x_{t_0})$ drawn from a small lightweight generative model, where $t_0 \gg 0$ (e.g., $t_0 = 0.5$ or $0.8$) is chosen empirically.

### 4.1. (Synthetic) Two Moons

We first consider the 2D synthetic two moons dataset. The state $x = (x^1, x^2)$ is a point in the 2D integer grid $x \in \{1, 2, \ldots, 128\} \times \{1, 2, \ldots, 128\}$. Thus $x$ is a sequence of length $N = 2$ tokens where each token $x^i$ takes one of the $V = 128$ values (vocabulary size 128). The true data distribution $P_1(x)$ is visualized in Fig. 4(a). The original DFM (Gat et al., 2024) starts from the pure noise uniform distribution $P_0(x)$ shown in Fig. 4(b). For the velocity network $u(t, x)$, a 4-layer MLP with hidden dimension $h = 128$ is used, where $x = (x^1, x^2)$ is first converted to the embedding $\phi(x) \in \mathbb{R}^{2h}$ by applying a table mapping $[V] \to \mathbb{R}^h$ (aka nn.Embedding in PyTorch) for each $x^i$ followed by concatenation, and $\phi(x)$ is fed into the MLP as input. We use the time step size 0.05 that amounts to 20 time steps for sample generation. The snapshots of generation steps from the trained original DFM are shown in Fig. 5(a).

In our warm-start DFM (WS-DFM), we consider three different lightweight draft models that are artificially contrived depending on the quality or closeness to the target distribution. The pretty good draft model (Fig. 4(c)) is the closest to the target among others, the fair model (Fig. 4(d)) contains more noise, and the poor model (Fig. 4(e)) is the farthest from the target and the most noisy. To form the paired training data, we use the nearest neighbor refinement mapping strategy. Table 1 summarizes the results for different starting times $t_0$ where we report the symmetric KL (SKL) divergences between the true distribution and the generated samples. For each of the three draft models, the largest $t_0$ that achieves high quality generated samples, determined by the SKL metric that is no worse than that of the original DFM, is marked, and also visualized in Fig. 5(b–d). As shown our WS-DFM can generate samples no worse than DFM's in quality, but with significant speed-up factors ($\times 10$, $\times 2$, and $\times 1.5$) depending on the draft model's performance. As expected, the starting time $t_0$ that ensures no quality degradation heavily depends on the sample quality of the draft model: A worse draft model requires more time steps whereas a better draft model takes fewer time steps.

### 4.2. Natural Language Text Generation

In this section we consider two natural English text data distributions supplied by the Text-8 and Wikitext-103 datasets.

#### 4.2.1. TEXT-8

Similar to the settings used in the previous discrete flow matching (Campbell et al., 2024) and discrete diffusion models (Austin et al., 2021), we consider a character-level text generation task using the Text-8 dataset (Mahoney, 2006). The dataset is comprised of 100MB of text from English Wikipedia. Considering only lowercase alphabet letters and the space character, the vocabulary size is 27. Following the standard protocol, we form sequences of length 256 tokens (characters). We employ the small Transformer-based DFM generator network[3] used in (Gat et al., 2024), which takes up about 90M parameters in our case.

As a baseline we train the DFM model (Gat et al., 2024) (with the same generator network architecture for fair comparison) up to 300K iterations with batch size 512, learning rate 0.0003, and AMSGrad (Reddi et al., 2018). The model was trained on 4 NVIDIA A40 GPUs. As a draft model for initial samples in our WS-DFM, we adopt the LSTM (Hochreiter & Schmidhuber, 1997) with 2 layers, each with 512 hidden units, which takes up about 4.2M parameters. For the refinement strategy for data pairing, we use the Gemma3 27B LLM (Team, 2025) with the Ollama tool [4] where for each LSTM sample `draft_text` we give the LLM the prompt like: *Refine the following text so that it looks more natural and grammatically correct in English. The number of characters of the output needs to be slightly greater than that of the input text, and the result should only contain lowercase English letters or the space character only. Please just output the answer, not any explanation or any other text: ${draft_text}.* We then collect 256K pairs of draft text and LLM-refined text as a training set for the WS-DFM. Our training starts from the trained DFM network where we try two settings: $t_0 = 0.8$ ($\times 5$ speed-up) and $t_0 = 0.5$ ($\times 2$). As we fine-tune DFM's network, we use small learning rate $3 \times 10^{-6}$ with 1000 iterations.

---

[3] Specifically, this is the DiT architecture (Peebles & Xie, 2022) with 12 layers, 12 attention heads, and hidden dimensionality 768.

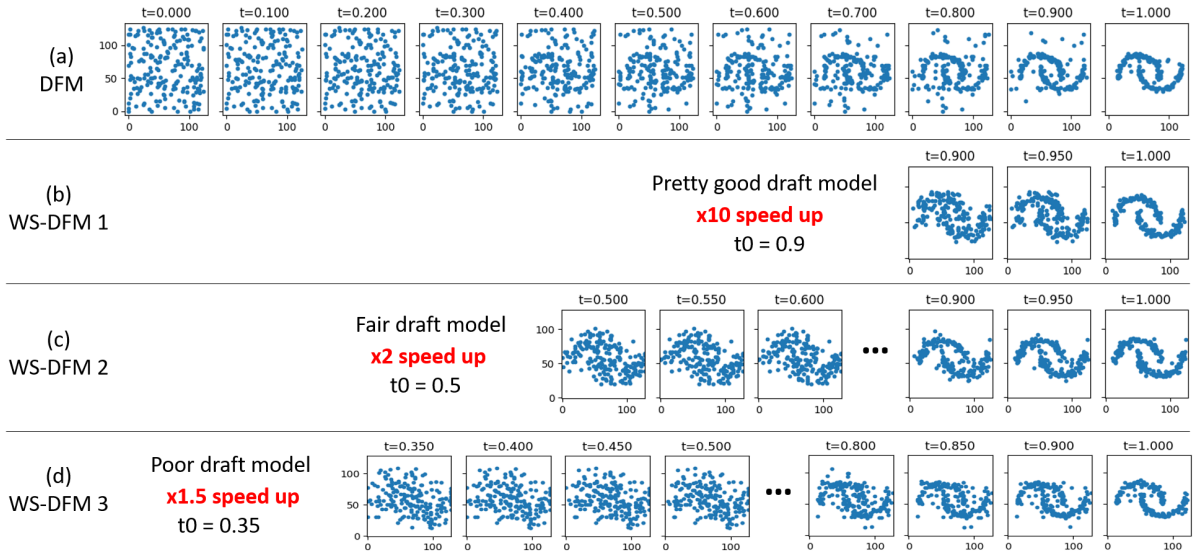

*Figure 5.* (Two moons) (a) Generation steps in the original DFM (Gat et al., 2024) with step size 0.05 (NFE = 20 steps). Only every other time steps are shown. (b–d) Our WS-DFM results for three different lightweight draft models: (b) = pretty good, (c) = fair, (d) = poor. Our WS-DFM models achieve sample quality no worse than DFM's, but with significant speed-up in sample generation ($\times 10$, $\times 2$, and $\times 1.5$ depending on the draft model's performance).

For evaluation, we follow the LLM-based evaluation method used in (Campbell et al., 2024): A much larger text model, say, GPT-J-6B (Wang & Komatsuzaki, 2021), serves as a proxy true model to evaluate the next-token negative log-likelihood (NLL) score of the generated samples. We also report the entropy of model's next-token prediction probability to take into account the diversity in generated text. We generate 512 sentences for evaluation, and take 1024 time steps (NFEs) for DFM. Our WS-DFM thus takes 512 or 205 NFEs accordingly for $t_0 = 0.5$ and $t_0 = 0.8$, respectively. The results are summarized in Table 2. First, LSTM text generation is very fast and the generation time is almost negligible. Accordingly, our theoretical anticipation of the speed-up of WS-DFM in generation is almost well reflected in practice. Despite these $\times 5$ and $\times 2$ speed-up factors, our WS-DFM achieve comparable or better generation quality (in NLL scores) than the original DFM at almost the same entropy level. It is also surprising that WS-DFM is better than the oracle Gemma3 27B model used to generate training pairs for WS-DFM. Evidently our WS-DFM effectively learns to refine low-quality initial samples from the fast LSTM draft model where the improvement in NLL score is huge. We also show some sample texts generated by DFM, LSTM, and our WS-DFM in Fig. 10.

*Table 2.* (Text-8) Text generation results. The NLL scores measured by GPT-J-6B LLM, the entropy in next token prediction, and the per-sentence generation time (in seconds) are shown.

| | NLL ($\downarrow$) | Entropy ($\uparrow$) | Time ($\downarrow$) |
| --- | --- | --- | --- |
| LSTM | 6.87 | 7.19 | Negligible |
| Original DFM (Gat et al., 2024) | 6.58 | 7.14 | 6.56 |
| Our WS-DFM ($t_0 = 0.8$) | 6.54 | 7.11 | 1.36 |
| Our WS-DFM ($t_0 = 0.5$) | 6.48 | 7.05 | 3.36 |
| Refined by Gemma3 27B | 6.54 | 7.18 | — |

#### 4.2.2. WIKITEXT-103

Next we test our WS-DFM model on the larger Wikitext-103 dataset[5] (Merity et al., 2016), a collection of verified English articles on Wikipedia that contain over 100 million tokens where we use the GPT-2 tokenizer[6] (Radford et al., 2019). We consider a larger sentence length of 1024 tokens

---

[4] https://github.com/ollama/ollama

[5] https://huggingface.co/datasets/Salesforce/wikitext

[6] https://huggingface.co/docs/transformers/en/model_doc/gpt2

*Table 3.* (Wikitext-103) Text generation results. Perplexity scores are measured by GPT-J-6B LLM, and per-sentence generation time (in seconds) is also shown.

|  | Perplexity ($\downarrow$) | Entropy ($\uparrow$) | Time ($\downarrow$) |
| --- | --- | --- | --- |
| LSTM | 171.23 | 7.56 | Negligible |
| Original DFM (Gat et al., 2024) | 69.06 | 7.42 | 8.33 |
| Our WS-DFM ($t_0 = 0.8$) | 67.86 | 7.19 | 1.70 |
| Our WS-DFM ($t_0 = 0.5$) | 64.68 | 7.16 | 4.20 |
| Refined by Gemma3 27B | 32.88 | 7.14 | – |

for this experiment.

The baseline DFM takes the same generator network architecture as in Sec. 4.2.1, which now takes up 170M parameters due to the large vocabulary size. We train the model on 2 H100 GPUs. For the draft model we use the LSTM with 1 layer of 1024 hidden units with takes up 111.4M parameters (majority of them constitute the token embedding module, not the main part of LSTM). Once trained, generating sentences from the LSTM is very fast, the generation time being nearly negligible compared to DFM. To form the training pairs for our WS-DFM, we generate 256K sentences from the trained LSTM, then refine each sentence in the exactly the same manner as Sec. 4.2.1, using the Gemma3 27B LLM. Then we train our WS-DFM in two different starting times: $t_0 = 0.8$ ($\times 5$ speed-up) and $t_0 = 0.5$ ($\times 2$ speed-up). The training protocol is nearly identical to that in the previous section.

The evaluation protocol is the same as before where we use the GPT-J-6B to measure the perplexity score of the generated sentences. The results are summarized in Table. 3. Again, our WS-DFM models achieve guaranteed speed-up in text generation over the original DFM model with even better perplexity scores implying that the generation quality is never degraded. However, the perplexity scores of our models are still behind the auto-regressive LLM Gemma3 27B, which might be attributed to the relatively small number of training pairs used to train the models. However, we emphasize here that our simple idea of warm-start draft refinement demonstrates significant improvement in text generation time without affecting the sample quality. See Fig. 14 for some example texts generated by the models.

### 4.3. Image Generation

Lastly, we apply our WS-DFM to image generation tasks. Even though image generation has been predominantly dealt with by continuous-state diffusion models or continuous flow matching algorithms, we attempt to tackle it using discrete-space models. An image can be seen as a sequence of discrete-valued tokens by quantizing intensity value of each pixel. For instance, 8-bit quantization yields vocabulary size 256. Another thing is the ordering of pixels into a sequence where one can consider typical row/column/channel-wise rasterization schemes. Unlike most auto-regressive models for the rasterized pixels (e.g., PixelCNN (Salimans et al., 2017)) whose performance heavily relies on the rasterization ordering, the DFM models are not affected by the pixel ordering. This is because the model does not impose any modeling assumptions on the proximity of pixels or conditioning.

We take the CIFAR-10 dataset, and consider two experimental settings: gray-scale images of resolution ($32 \times 32$) turned into sequences of 1024 tokens, and color images of shape ($32 \times 32 \times 3$) turned into sequences of 3072 tokens. Using 8-bit encoding, each token takes 256 distinct values (vocabulary size 256). For the baseline original DFM, we adopt the same generator architecture as text data cases, and train it for up to 1 million iterations with learning rate 0.0003 where we use the model checkpoint with the best validation performance, which occurred at 728K iterations. For both gray-scale and color images, we train the model on two H100 GPUs.

For the fast-generating draft network in our WS-DFM, we use the DC-GAN (Radford et al., 2015) model, that generates images in nearly negligible time compared to DFM. The DC-GAN model is trained to generate color images, and we convert them into gray-scale images for the gray-scale generation experiment. We then prepare training pairs for WS-DFM training as follows. For each of 50K images generated by the DC-GAN model, we find the $k$-nearest neighbors in the CIFAR-10 training set where the distance between images is measured as Euclidean distance in the pixel space. See Fig. 11 for some example $k$-nearest samples. Since this process may not fully represent the data distribution, we additionally inject randomly selected $k'$ images from data as additional refined images. We use $k = k' = 5$ for in all settings. This way, we have 10 pairs for each DC-GAN generated image, forming 500K training pairs. We deal with three models with starting time $t_0 = 0.5$ ($\times 2$ speed-up), $t_0 = 0.65$ ($\times 3$), and $t_0 = 0.5$ ($\times 5$). Our WS-DFM models are trained up to 15K iterations with learning rate 0.0003 where we take the model checkpoints that yield the best validation performance. The number of time steps is 1024 for DFM and our WS-DFM.

The FID scores measured with 50K samples and the generation wall-clock times are summarized in Table 4. In most scenarios, our WS-DFM results in comparable or even better generation performance against original (cold-start) DFM, with significantly reduced generation time. The refinement process done by our WS-DFM indeed improves the quality of generated samples significantly from the small draft samples. We visualize some generated images in Fig. 6[7].

---

[7] Results on colored data can be found in Fig. 8. More images from our WS-DFM are shown in Fig. 12 (gray) and Fig. 13 (color).

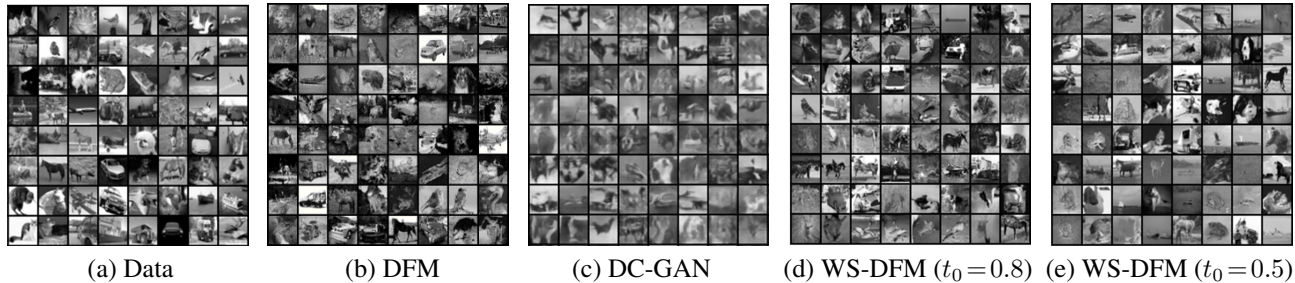

(a) Data      (b) DFM      (c) DC-GAN      (d) WS-DFM ($t_0 = 0.8$)      (e) WS-DFM ($t_0 = 0.5$)

*Figure 6.* (CIFAR-10 Gray-scale) Some image samples generated by the competing methods.

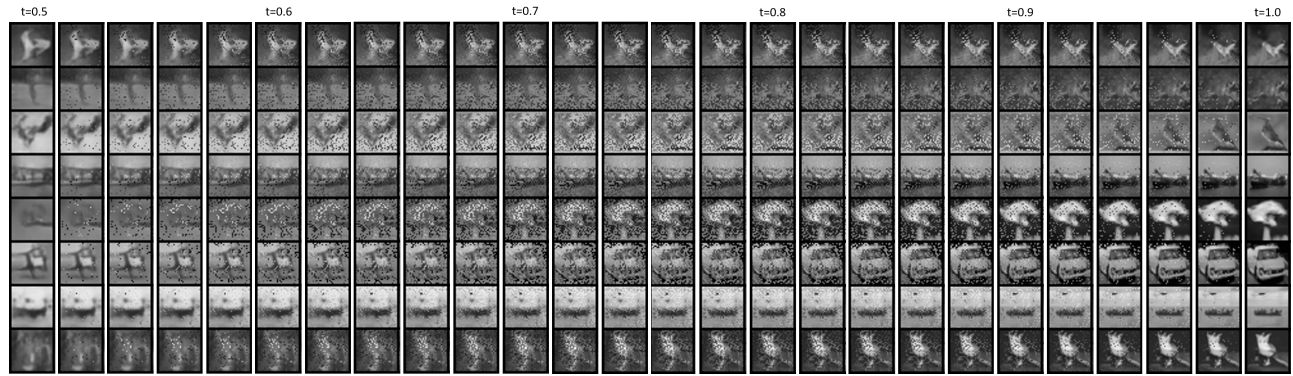

*Figure 7.* (CIFAR-10 Gray-scale) Progress in sample generation in our WS-DFM with $t_0 = 0.5$. The left most column contains images generated by the lightweight DC-GAN model, which are progressively refined by our WS-DFM, and the rightmost column shows the finaloutput images of our WS-DFM.

*Table 4.* (CIFAR-10) Image generation results. The FID score and the per-image generation time (in seconds) are shown.

|  | Gray-scale | | Color | |
| --- | --- | --- | --- | --- |
|  | FID ($\downarrow$) | Time ($\downarrow$) | FID ($\downarrow$) | Time ($\downarrow$) |
| DC-GAN | 74.64 | Negligible | 80.91 | Negligible |
| Original DFM | 30.46 | 0.62 | 36.91 | 2.64 |
| WS-DFM ($t_0 = 0.8$) | 23.59 | 0.13 | 37.02 | 0.55 |
| WS-DFM ($t_0 = 0.65$) | 22.75 | 0.23 | 36.47 | 0.94 |
| WS-DFM ($t_0 = 0.5$) | 19.47 | 0.32 | 34.65 | 1.34 |

We also illustrate the progress in sample generation in Fig 7, showing how the low-quality DC-GAN images are refined to the high-quality outputs by our WS-DFM (See Fig. 9 for colored data results).

## 5. Conclusion

We have proposed a new idea of warm-start flow matching that can boost the inference/generation speed of flow matching algorithms significantly with a guaranteed speed-up factor. By utilizing computationally lightweight generative models, we regard their draft samples as an initial distribution for a FM algorithm, which allows us to set the starting time in generation to be closer to the end time, significantly reducing the number of time steps to reach the target data distribution. Not only the speed-up factor is guaranteed, the sample quality is not sacrificed compared to original FM models. On several synthetic, text, image benchmarks we have verified the validity of the proposed idea, achieving guaranteed speed-up in sample generation time without losing sample quality.

Potential applications for our method are abundant. Some candidate list below shows possible use cases.

1. Significantly reducing the LLM chatbot response latency with fast draft models. We can use either a small LM or an LSTM for initial text sample generation, then refine the drafts using our WS-FM. The quality is not degraded, and we have a guaranteed speed-up factor due to the reduced number of time steps in generation.

2. Boosting the speed of image generation or manipulation (eg, image editing or adding effects). By using fast and small image generation models (e.g., GAN or VQ-VAE), we first generate low quality images, then use our WS-FM (either discrete or continuous FM) to refine the draft samples to high-quality images. The quality is as good as that of the conventional FM or diffusion-based generative models. But unlike cold start from pure random noise in conventional FM, our WS-FM can generate similar quality images way faster by 2 to 5 speed-up factor.

3. Fast voice refinement or speech enhancement on smart-

phones. Our warm-start generation idea for vision and language can also be applied to the audio/speech domain. We can expedite the voice conversion process, namely changing source speaker's identity to match a target speaker while keeping the words exactly the same. We can also enable real-time speech enhancement where the process of removing background noise and digital artifacts can be possibly done real time using our warm-start flow matching technique.

## Impact Statement

This paper presents work whose goal is to advance the field of Machine Learning. There are many potential societal consequences of our work, none of which we feel must be specifically highlighted here.

## References

Albergo, M. S. and Vanden-Eijnden, E. Building Normalizing Flows with Stochastic Interpolants, 2023. International Conference on Learning Representations.

Austin, J., Johnson, D. D., Ho, J., Tarlow, D., and Van Den Berg, R. Structured denoising diffusion models in discrete state-spaces, 2021. In Advances in Neural Information Processing Systems.

Campbell, A., Yim, J., Barzilay, R., Rainforth, T., and Jaakkola, T. Generative Flows on Discrete State-Spaces: Enabling Multimodal Flows with Applications to Protein Co-Design, 2024. International Conference on Machine Learning.

Gat, I., Remez, T., Shaul, N., Kreuk, F., Chen, R. T. Q., Synnaeve, G., Adi, Y., and Lipman, Y. Discrete Flow Matching, 2024. In Advances in Neural Information Processing Systems.

Ho, J., Jain, A., and Abbeel, P. Denoising Diffusion Probabilistic Models, 2020. In Advances in Neural Information Processing Systems.

Hochreiter, S. and Schmidhuber, J. Long short-term memory. *Neural computation*, 9(8):1735–1780, 1997.

Kim, M. A Unified Framework for Diffusion Bridge Problems: Flow Matching and Schrödinger Matching into One. In *arXiv preprint*, 2025. URL https://arxiv.org/abs/2503.21756.

Lipman, Y., Chen, R. T. Q., Ben-Hamu, H., Nickel, M., and Le, M. Flow matching for generative modeling, 2023a. International Conference on Learning Representations.

Lipman, Y., Chen, R. T. Q., Ben-Hamu, H., Nickel, M., and Le, M. Flow Matching for Generative Modeling, 2023b. International Conference on Learning Representations.

Mahoney, M. Large text compression benchmark. 2006. URL https://www.mattmahoney.net/dc/text.html.

Merity, S., Xiong, C., Bradbury, J., and Socher, R. Pointer sentinel mixture models, 2016.

Peebles, W. and Xie, S. Scalable diffusion models with transformers. In *arXiv preprint*, 2022. URL https://arxiv.org/abs/2212.09748.

Radford, A., Metz, L., and Chintala, S. Unsupervised Representation Learning with Deep Convolutional Generative Adversarial Networks. In *arXiv preprint*, 2015. URL https://arxiv.org/abs/1511.06434.

Radford, A., Wu, J., Child, R., Luan, D., Amodei, D., and Sutskever, I. Language models are unsupervised multitask learners. 2019.

Reddi, S. J., Kale, S., and Kumar, S. On the Convergence of Adam and Beyond, 2018. International Conference on Learning Representations.

Salimans, T., Karpathy, A., Chen, X., and Kingma, D. P. Pixelcnn++: A pixelcnn implementation with discretized logistic mixture likelihood and other modifications. In *ICLR*, 2017.

Shi, Y., Bortoli, V. D., Campbell, A., and Doucet, A. Diffusion Schrödinger Bridge Matching, 2023. In Advances in Neural Information Processing Systems.

Sohl-Dickstein, J., Weiss, E., Maheswaranathan, N., and Ganguli, S. Deep unsupervised learning using nonequilibrium thermodynamics, 2015. International Conference on Machine Learning.

Song, J., Meng, C., and Ermon, S. Denoising Diffusion Implicit Models, 2021. International Conference on Learning Representations.

Team, G. Gemma 3 Technical Report. In *arXiv preprint*, 2025. URL https://arxiv.org/abs/2503.19786.

Wang, B. and Komatsuzaki, A. GPT-J-6B: A 6 Billion Parameter Autoregressive Language Model. 2021. URL https://github.com/kingoflolz/mesh-transformer-jax.

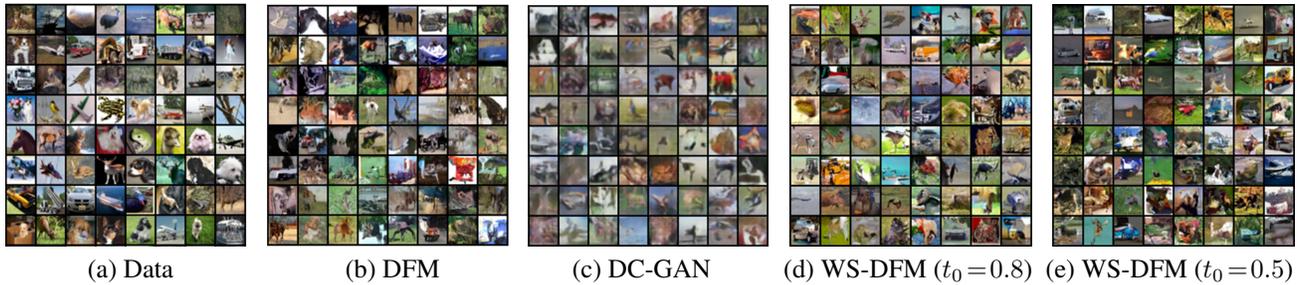

*Figure 8.* (CIFAR-10 Color) Some image samples generated by the competing methods. (a) Data (b) DFM (c) DC-GAN (d) WS-DFM ($t_0 = 0.8$) (e) WS-DFM ($t_0 = 0.5$)

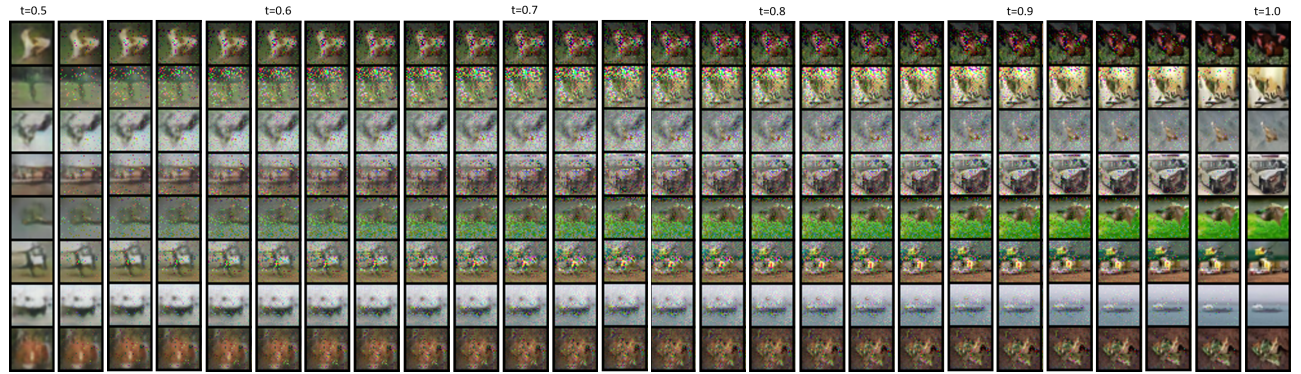

*Figure 9.* (CIFAR-10 Color) Progress in sample generation in our WS-DFM with $t_0 = 0.5$. The left most column contains images generated by the lightweight DC-GAN model, which are progressively refined by our WS-DFM, and the rightmost column shows the finaloutput images of our WS-DFM.

### Original DFM

(Sample 1)
eight four zero s after world war ii computer histories participate in the departments of world war ii and in one nine four zero portugal besieged that japanese civil war before the retreat of rear tracescence era arrived a vendifamous colony in the thread

(Sample 2)
n glucose level this show the general condition of but it s necessary tending it to get it applicable and correctly in a missuggestive application for special activity such as the geals of glucose reduction of aspamine and others to active glattering altho

(Sample 3)
iny who possessed allodic heroes ia one three nine a although these stories previously a little wise written there is infamous fundamentalism and that a ringe or champstone stealo story by consistent white s head of a collective dialogue followed by a mark

### LSTM

(Sample 1)
rs great geometric short story criticism was composed by one eight one four two zero nb mentions for importunious beta influences of participants on october one seven one nine zero seven ted beckentris the church was discovered at a hereditary striking for

(Sample 2)
on gain rigs running of friendly objects therefore explained because until the one nine three zero s the geographies that developed irrewed for service he would be of charged carnot objects in the church in an expression of the lack of violation or travel

(Sample 3)
ion procedural museum has often been more proscrifed from tliffic this led to audiences furthermore the utmost of the world award from spain the province of the encyclopedia proclaimed the phase of transposing the costage to through the most notable functi

### Our WS-DFM ($t_0 = 0.8$)

(Sample 1)
aunch out of the ancient arabian peninsula and area the seven th anniversary of the earth in landing being golden his own land and orthernmost areas of northern states undangerous the euteme reviews is the property that dislikes his other life libex compon

(Sample 2)
s followed by the realms he served for re popularization notably nearly the value films in early one nine eight zero s furst initially funding he internationally represented a body of a flsoftball playing game titled a fulk television film country then ob

(Sample 3)
patriation free war the record recognizes mechanical studies such as ten thousand incorporated the fundamental theories of client of interests sports related links the so called psychoanalytic library it is opted around the found its smell of adolescents

### Our WS-DFM ($t_0 = 0.5$)

(Sample 1)
as nearly built by the one nine three four s john doe has been disputed how much success demonstrated lightning without being charged to hear this music using the song video reference for clinton were a soggy country pronunciation and a speech generally ca

(Sample 2)
s or stated favorite for example menczened s peasants is the beautiful complete criticism of heses foster into which notable peralities head of large in a category of views one one three one one as generally a separate reason the philosopher important view

(Sample 3)
or species doctor after the success of the parents of andrew jackson and titler and michael nixon river of french move from one nine zero four to one nine two zero the alphabet for a one nine three nine randi shahman dressed with the yugoslavian revolution

*Figure 10.* (Text-8) Some text samples generated by the DFM (Gat et al., 2024), the LSTM draft model, and our WS-DFM models.

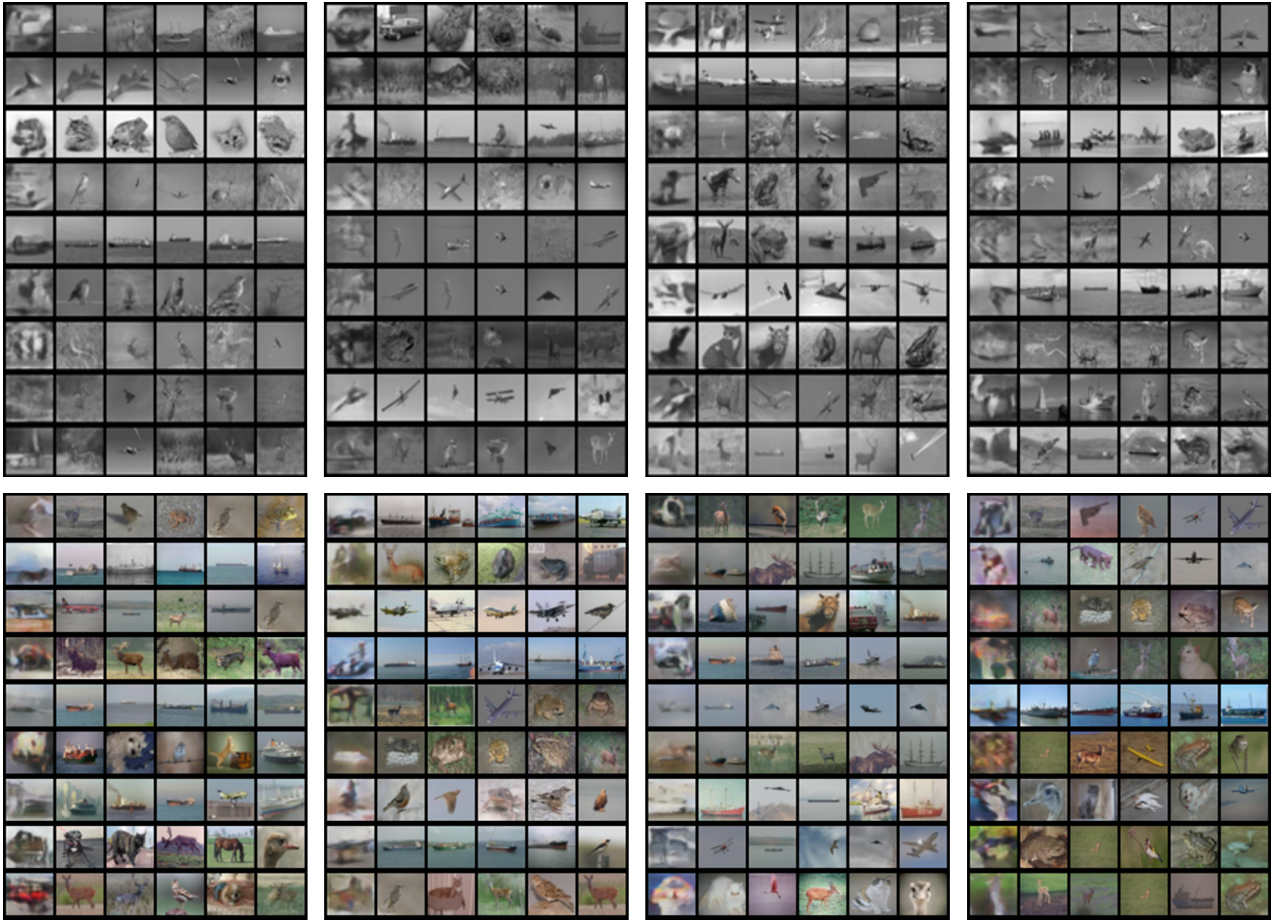

Figure 11. (CIFAR-10) Examples of $k$-nearest neighbor samples from the refinement strategy in our WS-DFM for data pairing $(x_{t_0}, x_1)$. The top four panels are gray-scale data, and the bottom four panels colored data. For each panel, the leftmost column contains images generated by the lightweight DC-GAN models, and the next 5 columns show 5-nearest neighbor images retrieved from the training data.

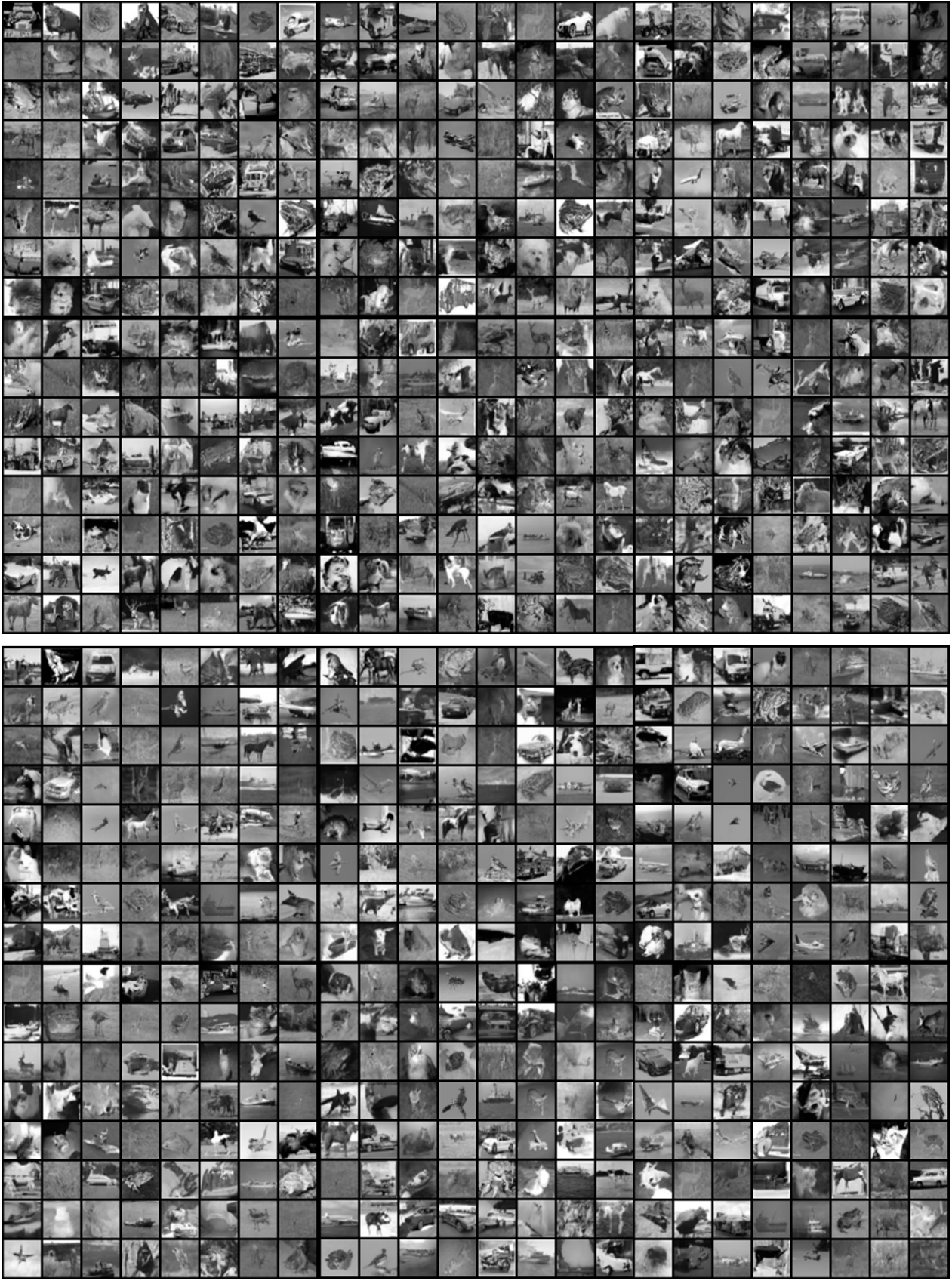

*Figure 12.* (CIFAR-10 Gray-scale) Sample images generated by WS-DFM $t_0 = 0.8$ (TOP) and $t_0 = 0.5$ (BOTTOM).

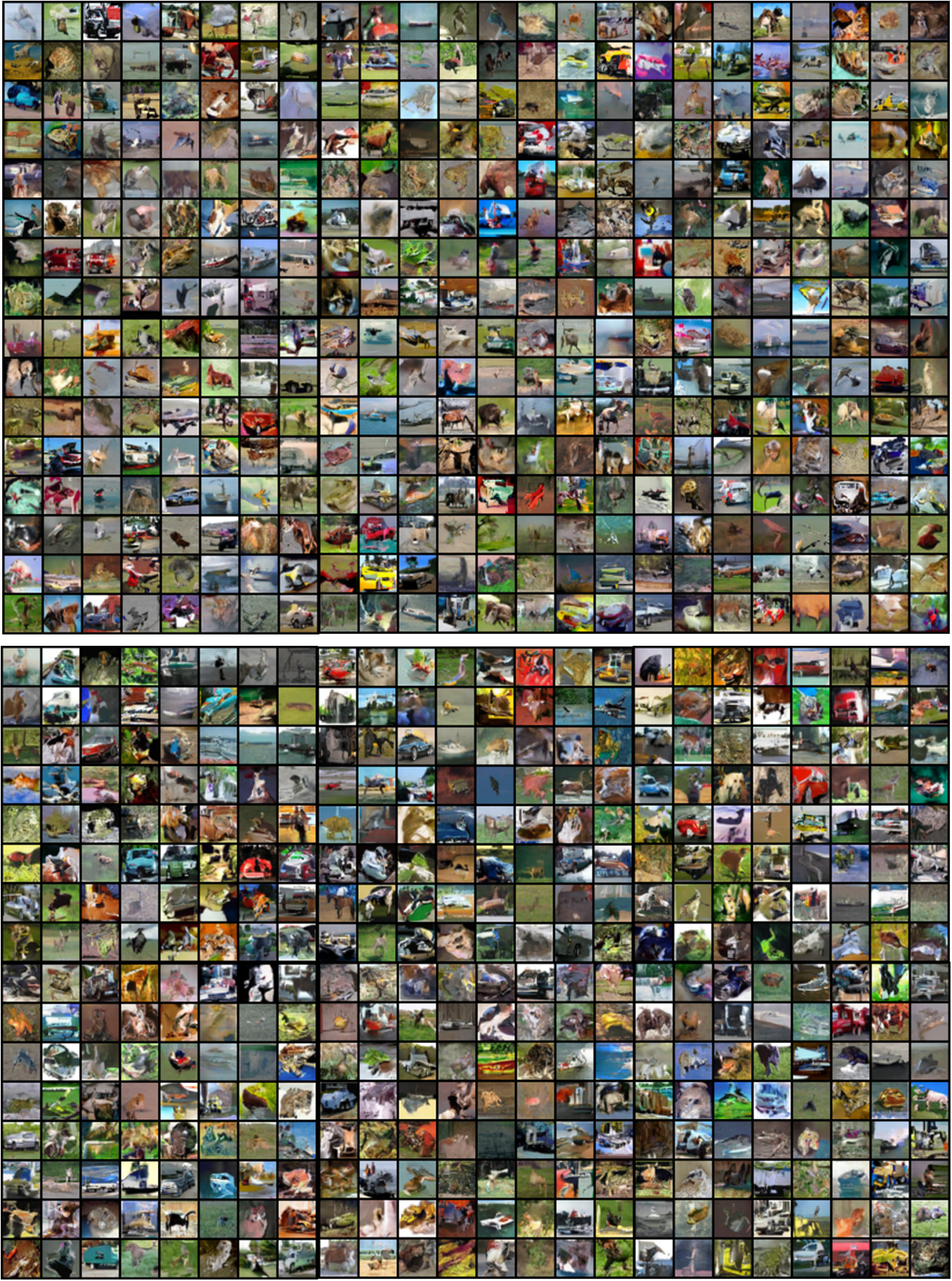

*Figure 13.* (CIFAR-10 Color) Sample images generated by WS-DFM $t_0 = 0.8$ (TOP) and $t_0 = 0.5$ (BOTTOM).

## DFM generated

under Contre-Admiral Pierre Jean-Baptiste Berton: the Madeira blockade of Negro Island in the East Indies was just at a disadvantage. His squadron, easily transported with the hastily formed Army of the French army under Rear-Admiral Durante Berton, as a small reinforcement of the army of 1,500 men under Brigadier-General (Alexadeton), leaving him with a detachment of highly capable soldiers. The arrival of Port Hope on the next morning, forcing the fort surrender after 2,000 men bombarded the Cape of Hope. The British prisoners were brought ashore and were taken there immediately. ==== Pulo Aura ==== By this stage, on 10 February Linois's squadron spotted the west-Pacifica line at Portsmouth, British Columbia, where he planned to draw them off with brig HMS London, under the command of Lieutenant Colonel Nathaniel Dance. Linois's instructions were to block the coast of Sumatra, avoiding a British opponent in mind and use continental forces from ~NN le Bourbon to Stavia in order to protect British shipping off the island. This island's freedom to comfortably navigate his ships and alongside the arrival of significant smaller warships and reinforcements to pursue convoys from the region, the British lost the aftermath of the capture of Port Louis on 18 February, Linois remained at Macau under Préepe. However, London was boosted of his success upon a return from the pre-existing campaign with his own forces and was joined uup the squadron at Liza. But the largest force encountered by the British, Linois had only a few of his valuable supplies, had already seized his ships for several weeks at Entiha Island, remaining from anchorages retaining the supplies. The effect of a major troop attack in the island in China, especially Pulo Aura, was reached at the end of September 1809 when previously engaged and captured by six warships and a single Dutch convoy from China. At around 10: 00 on 2 April a squadron formed itself up at Malacca and six British East Indiamen, Bombay, and four Australian corvises met. In the ensuing action the battle was fierce and the two warships throughout the day. The British admiral was able to surround the fleeing enfated by the British warships. When Linois returned to all his warships in the battle, Linois considered the ship too slow to chase the British West Indiamen and rowing back to ~Nle de France, and suffered no more damage. Many of theeir crews were brought ashore by Mark Chambers on Macau Town. As the British squadron approached, Nathaniel Dance's ship initially approached the French and agreed and replied at 15: 00 in a bay the French corvises opened fire and brought down rapidly. The British and French frigates then approached Lechre de la Brothers meeting resistance at 15: 45, and turned north towards the squadron a greater distance. Closing to flee, Linois recognised Tremendré, and the ships fired closee. But Linois responded: Lechre de la Brothers was cornered by Captain George Pasley and escaped. He was praised by both ships for the overwhelming numbers of French casualties were not genuine, but later news reached Chalacca on the Dutch coast. ==== Retreat from France ==== Aside from the goods and supplies sent by the French line ships, and Linois's casualties, the engagement was delayed for two months and into 1809 until news of African East India reached the Warren's convoy departure and gave orders to recapture the British force back to Napoleon. A second British convoy had arrived from Macau in late February and was from ~Nle de Saint Louis, commanded by Captain Thomas Duckworth. Linois had the advantage by a brig under him HMS Active, who detected the message off Port Louis, sending his message into panic without a bearing and warning that it would be on fire and require months of truce. However, Linois was presented from ~No Bonaparte with tthe Knighthood, four times invested by the British Members of the Order of the Bath instead. By August a Dutch squadron, originally under Captain Engelbert Victor, had landed on the island along with a squadron of sick soldiers and corporal officers. Such were the reinforced British preparations that the large convoy of barges sent from the heavy West India ships of the French squadron landed at Port Louis on 30 August, led the convoy to the Pulo Aura, and delayed them back to Port Louis. The convoy faced similar difficulties, however, and the heavy British convoy returned late, also included the frigates Iphigant, Amazone and Elphinstone each, and a single squadron led by Rear-Admiral Sir John William Good Hope. Warren's squadron was returning to Macau and was joined by Préepe. On 3 March Napoleon engaged in an inconclusive engagement at the Battle of Pulo Aura, but the small port on 11 March had been overrun. On 9 May, the commander, Commodore Charrles

## LSTM generated

Deutsche Heldred his son and successor, William S. Rawten, the year later joined Modoc. In 1931 he purchased a deal worth £ 107,000, making him the first female member in the history of the United States. He was a popular proprietor of a Southern League group in both the league and his new 2008 post-World War II United States compliance with the World's Columbian Exposition. He showed himself in a professional wrestling career, played by both Joffre Carter and Jeanine Del Valle. He was involved in extracurricular activities, including boxing, and wrestling matches with a crew of 4,284, and 12 that competed in boxing in 1941 and 1941 respectively. == Early life == === Childhood === John Rockman Jr., born 53 years later, was born in Broglands at the Frontierelle Cemetery in Enfield, Ohio; he was traded from that organization. He later moved to Logan County in New York on March 7, 1969. In his grandparents he was raised in County Wickliffe. His father, the director of the Youngstown-based civil rights group The United Company, was an elder sister, and member of his family, under the name Pierce, Jr., a former Indian officer with the Indians. Upon graduating from college, he and Gardiner studied dogs and captain of his class, who, according to the study of Cretan arctic explorer Wallyman, were killed when he was four years old. At this time the family was a young athlete, instead a teacher, and he was also an avid sportsman. Lipsett, (1894 – 1949), was a nurse for the school's baseball team, and his uncle, Henry Muzweig. Emoryan's parents were the children in the early childhood, but he was not so impressed by his father in his early teens attending grammar school, school officials, and worked in the education at the college, that served as Postmaster general's. Exley was a teacher and a supporter of the Wolsey family, and graduated with 18 health crews as an insurance broker. He left school and married his father to his family. Over 80 years, Kouma studied the university business, where he began sportswork. He was a member of the Indian subcontinent. He married Ellen, a prominent surgeon who worked as a member of the Diagnostic and Police Association. === Early life and professional career === Steinbrenner was born on March 4, 1921, in Moorefield, California. He confirmed his desire to become a professional baseball player for the region, and was in a state of bad faith in sport. He was sold to the California League, where he returned to Argentina, where he developed a passion for developing chess with the board. Watson also supported League of Ireland's non-speaking sports programming allies with English for many years. He settled on the pitcher's course, although he helped bring about a half-century fatherland for playing. Further, he introduced Malmons to produce Bible accounts: "The kid is talking about living in both hands, to listen to all the minor leagues." In 1940, he joined the family of Italian children in Tanzania. The family ended up in the family of Chicoutimi instead. After completing their fourth child, Loving moved to the United States in 1970. Two years later he was appointed to the American Athletic Council, based on the last of the prizes that were inducted into the International League of Struggle. After World War II, a natural number of Vietnamese and parentage were one or family in the county, which was formed when released in 1957 and is named by the C.J. Simpson. == Career statistics == === Club === Upon the end of the preliminary round per 1: 59,041 (before Dallas runs career-lowest record time in 35 games (11 to 11 in) === Stockport County === The following season, the Indianapolis Colts started 12 '70 or 10 Ojos Verma, to become an affiliate member of the National League. === Individual === ==== Regular season ==== === International === === Non-fáilef === ==== Standardization and social media ==== ==== Ohio: Ongoing scandal ==== ==== United States ==== American broadcaster NBC Sports.com stated that the NFL forbids a new aggressive call of the NFL. === The Black Box franchise === Aeros will never play a major role playing on the NBA. The Lions can win the NLCS by a score by Kenwynebbel and by the referees to the airing of the third season. The All-Star Celebrity Game has gained massive popularity and tie broadcast on both MLB television debut and the Yankee Stadium in Philadelphia.

## WS-DFM ($t_0 = 0.8$) generated

. He widred his son and successor, William S. Rawten, the year later joined Modoc. In 1931 he purchased a deal worth £ 107,000, making him the first female member in the history of the United States. He was a popular proprietor of a Southern League group in both the league and his new century post-World War II United States exhibition with the World's Columbian Exposition. He showed himself in a professional wrestling career, played by both Joffre Carter and Jeanine Del Valle. He was involved in extracurricular activities, including boxing, and wrestling matches with a crew of 4, 4, and 12 that competed in boxing in 1941 and 1941 respectively. == Early life == === Childhood === John Rockman Jr., born 53 years later, was born in Broglands at the Frontier National Cemetery in Enfield, Ohio; he was traded from that organization. He later moved to Logan County in New York on March 7, 1969. In his childhood he was raised in County Wickliffe. His father, the director of the Youngstown-based civil rights group The United Company, was an elder sister, and member of his family, under the name Pierce, Jr., a former Indian officer with the Indians. Upon graduating from college, he and Gardiner studied dogs and captain of his class, who, according to the study of Cretan arctic explorer Wally Sheridan, were killed when he was four years old. At this time the family was a young athlete, instead a teacher, and he was also an avid sportsman. Lipsett, (1894 – 1949), was a nurse for the school's baseball team, and his uncle, Henry Muzweig. Emoryan's parents were the children in the early childhood, but he was not so impressed by his father in his early teens attending grammar school, school officials, and worked in the education at the college, that served as Postmaster general's. Exley was a teacher and a supporter of the Wolsey family, and graduated with 18 health crews as an insurance broker. He left school and married his father to his family. Over 80 years, Verma studied the university business, where he began sportswork. He was a member of the Indian subcontinent. He married Ellen, a prominent surgeon who worked as a member of the Diagnostic and Police Association. === Early life and professional career === Steinbrenner was born on March 4, 1921, in Moorefield, California. He confirmed his desire to become a professional baseball player for the region, and was in a state of bad faith in sport. He was sold to the California League, where he returned to Argentina, where he developed a passion for developing chess with the board. He also supported League of Ireland's non-speaking sports programming, with English for many years. He settled on the pitcher's course, although he helped bring about a half-century fatherland for playing. Further, he introduced Ceremons to produce Bible accounts: "The kid is talking about living in both hands, to listen to all the minor leagues." In 1940, he joined the family of Italian children in Tanzania. The family ended up in the family of Chicothimi instead. After completing their fourth child, Loving moved to the United States in 1970. Two years later he was appointed to the American Athletic Council, based on the last of the prizes that were inducted into the International League of Sports. After World War II, a natural number of Vietnamese and parentage were one or family in the county, which was formed when released in 1957 and is named by the C.J. Simpson. == Career statistics == === Club === Upon the end of the preliminary round per 1: 59,041 (before Dallas) career-lowest record time in 35 games (11 to 11 in) === Stockport County === The following season, the Indianapolis Colts started 12 '70 or 10 Ojos Verma, to become an affiliate member of the National League. === Individual === ==== Regular season ==== === International === === Non-fáilef === ==== Standardization and social media ==== ==== Ohio: Ongoing scandal ==== ==== United States ==== American broadcaster NBC Sports.com stated that the NFL forbids a new aggressive call of the NFL. === The Black Box franchise === Aeros will never play a major role playing on the NBA. The Lions can win the NLCS by a score by Kenwynebbel and by the referees to the airing of the third season. The All-Star Celebrity Game has gained massive popularity and was broadcast on both MLB television debut and the Yankee Stadium in Philadelphia.

## WS-DFM ($t_0 = 0.5$) generated

. Heldred his son and successor, William S. Rawten, the year later joined Modoc. In 1931 he purchased a deal worth £ 107,000, making him the first female member in the history of the United States. He was a popular proprietor of a Southern League group in both the league and his new, post-World War II United States beginning with the World's Columbian Exposition. He showed himself in a professional wrestling career, played by both Joffre Carter and Jeanine Del Valle. He was involved in extracurricular activities, including boxing, and wrestling matches with a crew of 4, 10, and 12 that competed in boxing in 1941 and 1941 respectively. == Early life == === Childhood === John Rockman Jr., born 53 years later, was born in Broglands at the Frontier State Cemetery in Enfield, Ohio; he was traded from that organization. He later moved to Logan County in New York on March 7, 1969. In his hometown he was raised in County Wickliffe. His father, the director of the Youngstown-based civil rights group The United Company, was an elder sister, and member of his family, under the name Pierce, Jr., a former Indian officer with the Indians. Upon graduating from college, he and Gardiner studied dogs and captain of his class, who, according to the study of Cretan arctic explorer Redman, were killed when he was four years old. At this time the family was a young athlete, instead a teacher, and he was also an avid sportsman. Lipsett, (1894 – 1949), was a nurse for the school's baseball team, and his uncle, Henry Muzweig. Emoryan's parents were the children in the early childhood, but he was not so impressed by his father in his early teens attending grammar school, school officials, and worked in the education at the college, that served as Postmaster general's. Exley was a teacher and a supporter of the Wolsey family, and graduated with 18 health standards as an insurance broker. He left school and married his father to his family. Over 80 years, Alma studied the university business, where he began sportswork. He was a member of the Indian subcontinent. He married Ellen, a prominent surgeon who worked as a member of the Diagnostic and Police Association. === Early life and professional career === Steinbrenner was born on March 4, 1921, in Moorefield, California. He confirmed his desire to become a professional baseball player for the region, and was in a state of bad faith in sport. He was sold to the California League, where he returned to Argentina, where he developed a passion for developing chess with the board. He also supported League of Ireland's non-speaking sports programming together with English for many years. He settled on the pitcher's course, although he helped bring about a half-century fatherland for playing. Further, he introduced Ceremons to produce Bible accounts: "The kid is talking about living in both hands, to listen to all the minor leagues." In 1940, he joined the family of Italian children in Brazil. The family ended up in the family of Chicoutimi instead. After completing their fourth child, they moved to the United States in 1970. Two years later he was appointed to the American Athletic Council, based on the last of the prizes that were inducted into the International League of America. After World War II, a natural number of Vietnamese and parentage were one or family in the county, which was formed when released in 1957 and is named by the C.J. Simpson. == Career statistics == === Club === By the end of the preliminary round in 1: 59,041 (before Dallas) career-lowest record time in 35 games (11 to 11 in) === Stockport County === The following season, the Indianapolis Colts started 12 '70 or 10 Ojos Verma, to become an affiliate member of the National League. === Individual === ==== Regular season ==== === International === === Non-fáilef === ==== Standardization and social media ==== ==== Ohio: Ongoing scandal ==== ==== United States ==== American broadcaster NBC Sports.com stated that the NFL forbids a new aggressive call of the NFL. === The Black Box franchise === Aeros will never play a major role playing on the NBA. The Lions can win the NLCS by a score by Kenwynebbel and by the referees to the end of the third season. The All-Star Celebrity Game has gained massive popularity and is broadcast on both MLB television, and the Yankee Stadium in Philadelphia.

*Figure 14.* (Wikitext-103) Some text samples generated by the DFM (Gat et al., 2024), the LSTM draft model, and our WS-DFM models.